# Enhanced Biologically Inspired Model for Image Recognition Based on a Novel Patch Selection Method with Moment

Yan-Feng Lu[1], Li-Hao Jia [1], Hong Qaio[2, 3], Yi Li [4]

[1] Research Center for Brain-inspired Intelligence, Institute of Automation,

Chinese Academy of Sciences, Beijing, China

yanfeng.lv@ia.ac.cn

lihao.jia@ia.ac.cn

[2] The State Key Laboratory of Management and Control for Complex Systems, Institute of

Automation, Chinese Academy of Sciences, Beijing, China

[3] CAS Center for Excellence in Brain Science and Intelligence Technology, Shanghai, China

hong.qiao@ia.ac.cn

[4] School of Information Engineering, Nanchang University, Nanchang, China

littlepear@ncu.edu.cn

*Abstract* — Biologically inspired model (BIM) for image recognition is a robust computational architecture, which has attracted widespread attention. BIM can be described as a four-layer structure based on the mechanisms of the visual cortex. Although the performance of BIM for image recognition is robust, it takes the randomly selected ways for the patch selection, which is sightless, and results in heavy computing burden. To address this issue, we propose a novel patch selection method with oriented Gaussian-Hermite moment (PSGHM), and we enhanced the BIM based on the proposed PSGHM, named as PBIM. In contrast to the conventional BIM which adopts the random method to select patches within the feature representation layers processed by multi-scale Gabor filter banks, the proposed PBIM takes the PSGHM way to extract a small number of representation features while offering promising distinctiveness. To show the effectiveness of the proposed PBIM, experimental studies on object categorization are conducted on the CalTech05, TU Darmstadt (TUD), and GRAZ01 databases. Experimental results demonstrate that the performance of PBIM is a significant improvement on that of the conventional BIM.

## 1. Introduction

Image recognition has been widely applied in the applications of computer vision, such as robot navigation, pedestrian detection, and clinical diagnosis [1-3]. In the practical applications, the difficulties that arise in the image recognition are typically caused by variations in the appearance of the objects and the background complexity of the input images. The scale, rotation, and illumination variability, especially in the cluttered backgrounds, disturb the recognition performance strongly. For instance, various human postures (e.g., squatting, stooping, running, or standing) in a real environment make accurate recognition a difficult task. To address this issue, lots of methods have been proposed in the past years.

Conventional appearance based methods often use the global low-level visual features, e.g., gray value, color, border, and texture [4]. These methods usually take the extracted features into account equably; they do not selectively put particular emphasis on local discriminative features. Moreover, they are sensitive to occlusion, scale, illumination deformations. Local features based methods mix local descriptors and key point detectors with spatial information. The representational methods, e.g., scale-invariant feature transform (SIFT) [5], gradient location and orientation histogram (GLOH) [6], histogram of gradients (HOG) [7], and speeded up robust features (SURF) [8] have been proposed. Although these methods are effective in representing local discriminative features, they lack directional information. Even though bag-of-words (BoW) [9] and bag-of-features [10] are effective for resolving this issue; the amount of structure information still falls short.

Recently, significant advances have been made in the research of brain science [11-14]. The findings in the primary visual cortex V1 area are of significance. While researching the V1 area, Hubel & Wiesel discovered that the visual cortex analyzes features into various ways with different spatial orientations and frequencies [11]. The discovery gives an important support to early neuroscience theories. Based on these theories, Riesenhuber & Poggio described an original calculation framework for object recognition, called biologically inspired model (BIM) that tends to model the cognitive mechanism of the visual cortex [14]. Serre et al. upgraded the original BIM model and presented the standard BIM [15], which shows that the visual framework significantly improve the performance of object recognition. Lu et al. proposed a novel receptive field in the S1 layers and upgraded the framework by novel patch selection and matching processes [16-19]. Qiao et al. developed a modified BIM model, and employed it in a robot system [20-21]. In conclusion, these mentioned approaches get remarkable performance by fusing certain biologically motivated mechanisms.

The traditional BIM model uses patches that are randomly selected in the second (C1) layers, which generates a huge amount of redundant information and also prevents robustness against rotational deformation. The stored patches in the C1 layers are the key components of the discriminative and robust abilities of BIM. Superior features extracted by the stored patches determine the feature invariance and selectivity, preserving BIM performance in the cases of object appearance variation and cluttered backgrounds. The majority of patches selected by the random method, however, are redundant and not discriminative for the recognition task, which results in performance degradation and high computational cost. These drawbacks seriously limit the overall performance of BIM. We propose a solution to this issue, a novel patch selection method with oriented Gaussian-Hermite moment called PSGHM. In the PSGHM, we employ the oriented Gaussian-Hermite moment to represent the first layers (S1) of the BIM [16], and then the multi-scale keypoints are employed to locate the key regions of the S1, which aims to reduce the number of patches chosen but keep those with better discrimination than those chosen by random selection. We further propose a PSGHM-based BIM model (PBIM). We show its effectiveness, by applying it to object categorization and by conducting experimental studies on the CalTech05, TU Darmstadt (TUD), and GRAZ01 databases.

The remaining part of the article is organized as follows: in Section 2, we give an introduction about the conventional BIM; In Section 3, we describe the PSGHM method and PBIM model. In Section 4, we present experimental results based on three public databases. Finally, in Section 5, we give our conclusions.

## 2. BIM review

Conventional BIM is a computational framework with four layers: S1, C1, S2, and C2, which follows the mechanisms of the primary visual cortex and builds feature representation by patch matching and maximum pooling operations [15].

*S1* **layers**: The units in the S1 layers correspond to simple cells in V1. The S1 units take the form of Gabor functions [17], that model cortical simple cell receptive fields. Gabor functions are defined as

$$G(x, y) = \exp(-\frac{x_o^2 + \gamma^2 y_o^2}{2\sigma^2}) \times \cos(\frac{2\pi x_o}{\lambda}), \qquad (1)$$

$$s.t \quad x_o = x\cos\theta + y\sin\theta \quad \text{and} \quad y_o = -x\sin\theta + y\cos\theta,$$

where $\theta$ represents orientation, $\lambda$ is wavelength, $\sigma$ is scale, and $\gamma$ indicates spatial aspect ratio.

Given an input image, the S1 layer with orientation $\theta$ and scale $\sigma$ is calculated by

$$S1_{\sigma,\,\theta} = |G_{\sigma,\,\theta} * I|, \qquad (2)$$

where $*$ denotes convolution, $I$ is the input image, and $G_{\sigma,\,\theta}$ is a Gabor function with specific parameters.

*C1* **layers**: These layers describe the complex cells in V1. The layers are the dimensionally reduced S1 layers obtained by selecting the maximum over local spatial neighborhoods. This maximum pooling operation over local neighborhoods increases invariance (providing some robustness to shift and scale transformations).

*S2* **layers**: In these layers, S2 units pool over afferent C1 units from a local spatial neighborhood across all four orientations. The S2 layers describe the similarity between the C1 layers and stored patches in a Gaussian-like way using Euclidean distance. The responses of the corresponding S2 layers are calculated by

$$S2 = \exp(-\beta \| C1(j, k) - P_i \|^2 ), \tag{3}$$

where $\beta$ is the sharpness of the exponential function, $C1(j, k)$ denotes the afferent C1 layer with scale $j$ and orientation $k$, and $P_i$ is the $i_{th}$ patch from the previous C1 layers.

*C2* **layers**: The final set of shift- and scale-invariant C2 responses is computed by taking a global maximum of afferent S2 units across all scales and positions. The responses of the C2 layers are calculated by

$$C2 = \max_{(m,n,\sigma)}(S2(m,n,\sigma)). \tag{4}$$

where $(m, n)$ is the position of S2 units and $\sigma$ denotes the corresponding scale. The output is a vector of N C2 values, where N corresponds to the number of patches. The vector is used as the C2 feature in the recognition task. In contrast to the conventional BIM that the PBIM takes the PSGHM way to refine the representation features while offering promising distinctiveness.

## 3. Enhanced BIM model based on PSGHM (PBIM)

The BIM model is an appearance based descriptor that focuses on the invariance and selectivity of extracted features. Although conventional BIM is more flexible than some relevant descriptors [7-8] and its recognition performance is robust, it brings in huge numbers of redundant features by the random way, and results in heavy computing burden. In addition, the Gabor model has a high level of error in matching experimental physiological data [22]. To improve the performance of BIM, we proposed a novel patch selection way on oriented Gaussian-Hermite moment (PSGHM), and we enhanced the BIM model by the PSGHM to refine the representational features, named as PBIM.

The stored patches in the C1 layers are the key components of the discriminative and robust abilities of BIM; thus, the construction of a proper patch set is very important for the visual recognition task. A random selection of patches from the universal training set is an option, but that option is prone to bringing in huge amounts of redundant information and is sensitive to rotation. We address this by proposing a novel patch selection method based on Gaussian-Hermite Moment (PSGHM), which is based on a saliency mechanism and multi-scale keypoints on the OGHM layers. OGHM is a modified Gaussian-Hermite moment (GHM), whose properties are effective against scale change, image rotation, and illumination change [16], [23].

PSGHM consists of the following three steps: 1) Processing layer extraction, 2) salient regions construction, 3) keypoint candidate localization.

## 3.1 Processing layers

Input images are processed by OGHM filters with different directions and scales. We obtain OGHM scale pyramids as per the method in [16]. We make the directional multi-scale information tractable, by considering four orientations and sixteen scales for further processing in BIM. The processing layers can be calculated by:

$$N^{\theta}_{p,q}(i,j) = \frac{4}{(m-1)(n-1)} \sum_{i=0}^{m-1} \sum_{j=0}^{n-1} I(i+u-m/2+1, j+v-n/2+1) \times H^{\theta}_{p}(X,\sigma) H^{\theta}_{q}(Y,\sigma).$$

Where a digital image $I(i,j)$ whose size is $w \times h$ $[0 \leq i \leq w-1, 0 \leq j \leq h-1]$, $t(u,v)$ be a mask whose size is $m \times n$ $[0 \leq u \leq m-1, 0 \leq v \leq n-1]$, the oriented Gaussian-Hermite functions of the mask can be written as

$$\begin{cases} H^{\theta}_{p}(X,\sigma) = \frac{2}{m-1} \frac{1}{\sqrt{2^{p} p! \sqrt{\pi}\sigma}} \exp(-X^2/2\sigma^2) H_{p}(X/\sigma) \\ \\ H^{\theta}_{q}(Y,\sigma) = \frac{2}{n-1} \frac{1}{\sqrt{2^{q} q! \sqrt{\pi}\sigma}} \exp(-Y^2/2\sigma^2) H_{q}(Y/\sigma) \end{cases}$$

$H_{p}(X/\sigma)$ and $H_{q}(Y/\sigma)$ is the Gaussian-Hermite function on X and Y; the rotation variables X and Y can be calculated by

$$\begin{cases} X = x\cos\theta + y\sin\theta, \\ Y = -x\sin\theta + y\cos\theta. \end{cases}$$

## 3.2 Salient regions

A huge amount of irrelevant information exists in the processing layers, which complicates locating the more discriminative regions in the whole image. Obtaining dense distinctive features requires the construction of a salient region with rich discriminative information. Based on a biological visual perception mechanism, attention is an important visual processing stage that guides the gaze towards objects of interest in a visual scene [23]. This ability to orientate towards salient objects in a cluttered visual environment is of great significance because it allows rapid and accurate detection and tracking of prey or predators by organisms in the visual world. Itti and Koch first introduced a biologically inspired model to generate a saliency map [24]. In our paper, the saliency map is constructed in the processing layers based on a simple saliency model in [25].

The saliency map of the input image can be calculated:

$$Sal(I) = \left| (F^{-1}(e^{R(f)+i \cdot P(f)}))^2 \right|. \tag{8}$$

where $F$ is the Fourier transform, $f$ is frequency, $R(f)$ is the spectral residual, $P(f)$ is the phase spectrum of the image. More details can refer to [25].

We inhibit non-dominant information by adopting a simple version of the saliency map. We segment the constructed saliency map to obtain the salient region, i.e., where the distinctive features and patch extraction areas are concentrated. Given the saliency map of the input image, the salient region at

location $(x, y)$ can be obtained:

$$SR(x, y) = \begin{cases} 1 & if \ Sal(I(x, y)) > threshold, \\ 0 & otherwise. \end{cases} \quad (9)$$

In general, we set $threshold = M(Sal(I)) \times 2$, where $M(Sal(I))$ is the mean value of every pixel in the saliency map. (PSGHM experimentally shows the best performance when $threshold = M(Sal(I)) \times 2$, therefore we chose this value). The construction of the salient region is illustrated in Fig.1.

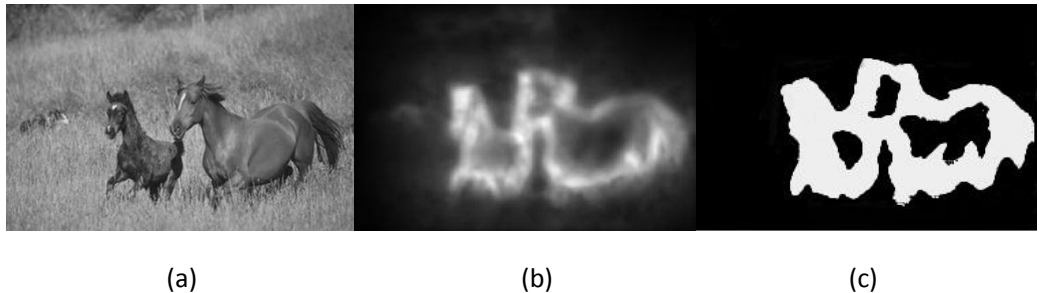

(a)          (b)          (c)

**Fig.1.** Construction of the salient region: (a) original image, (b) saliency map of the input image (c) salient region.

### 3.3 Keypoint candidate localization

In the constructed salient regions, we locate the keypoint candidates in each layer with their corresponding direction. In the conventional BIM model, patches are randomly extracted from the overall C1 layers to form the vocabulary of visual features. However, these visual features are neither refined nor discriminative; they include irrelevant and redundant information and degrade performance.

Achieving a reasonable recognition performance with BIM requires matching many patches, which results in high computational cost. PSGHM locates the keypoint candidates within the salient region, which are identified by a keypoint detection method named FAST [26]. FAST is widely used because of its accuracy and speed; however, it does not have an orientation component nor does it produce multi-scale features. Hence, we employ processing layers that are processed by Gabor scale pyramids at certain angles and produce FAST keypoints at each layer. In this way, we extract multi-scale keypoint candidates with specific angles. The keypoint candidate position $key$ can be localized by

$$key = FAST(P_{\theta,\sigma}(x, y)), \ (x, y) \in SR. \tag{10}$$

Here, $FAST$ is the keypoint detection method, $P_{\theta,\sigma}$ denotes the processing layer with orientation $\theta$ and scale $\sigma$, $(x, y)$ are pixel coordinates in the layer, and $SR$ is the salient region. We preferentially extract image patches around these detected keypoint candidates.

## 4. Experiments

We evaluate the performance of PBIM in several recognition tasks. In Section 4.1, we give the experiment setup. In Section 4.2, we evaluate the PBIM model under conditions of under normal circumstances using three datasets (Caltech5, TUD and GRAZ01).

## 4.1 Experiment Setup

Given the various appearance transformation of the images, we applied the position-scale-invariant C2 features of PBIM, and passed the features to a classifier to execute classification. (In the experiments of this article, we select the linear Lib-SVM [27] as the classifier). The other layers of PBIM are similar to those of the standard BIM, except for the obtained OGHM-based features in the $S1$ layers. We chose the evaluation metrics classification rate, recall, and 1-precision

$$1 - precision = \frac{number\ of\ false\ positives}{total\ number\ of\ positives}, \tag{23}$$

$$recall = \frac{number\ of\ true\ positives}{number\ of\ true\ positives\ and\ false\ negatives}, \tag{24}$$

$$classification\ rate = \frac{number\ of\ true\ positives\ and\ true\ negatives}{total\ number\ of\ positives\ and\ negatives}, \tag{25}$$

where a true positive is a correct classification of a positive (an object or scene), a true negative is a correct classification of a negative (background), a false positive is an incorrect positive classification and a false negative is an incorrect negative classification.

## 4.2 Experiment Evaluation

To evaluate the performance of PBIM, we compared it with that of other related algorithms on three public image datasets: CalTech5 [15]，TUD [28], and GRAZ01 [29].

### 4.2.1 Caltech5

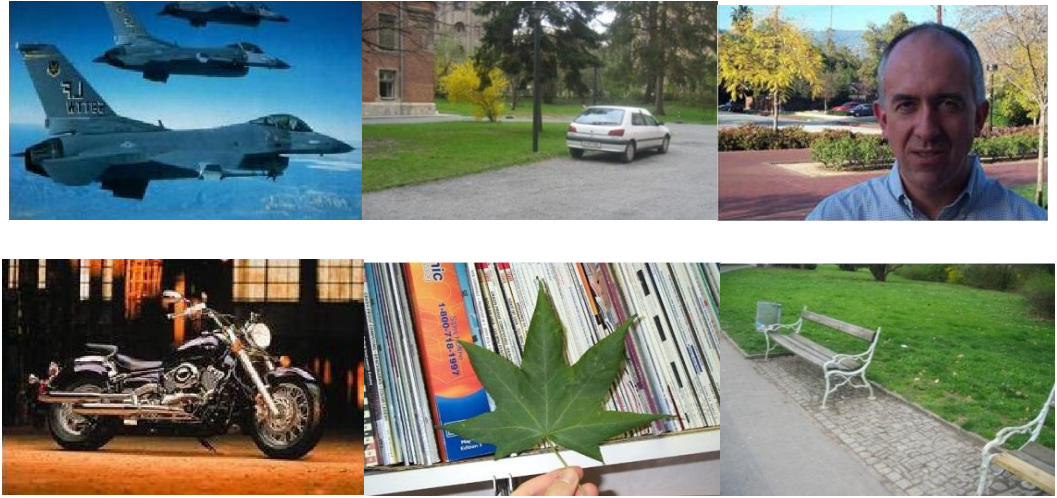

**Fig. 2.** Sampling images of the CalTech5 dataset. The last image is a background image.

The CalTech5 dataset contains the cars, frontal faces, aeroplanes, leaves, and motorcycles, as shown in Fig. 2. We applied this database to evaluate BIHM and make comparisons with the conventional BIM and the SIFT algorithm [5].

To make the experiment at a feature level and ensure a fair comparison between the methods, we compared the scale and position-invariant *C2* features produced by the standard BIM, and PBIM with SIFT features by passing the features to an SVM, which was trained to perform the object present/absent recognition task. We chose the classification rate for various numbers of features as the evaluation criterion. In the experiment, we randomly chose $15$ images from each category of the CaltTech5 dataset as positive training images and $15$ images from backgrounds as the negative training set. For the tests, $50$ other images (each category of the CaltTech5 dataset) and 50 other images (backgrounds) was randomly chosen as a testing set. It should be noted that a different number of features were randomly chosen from the *C2* layers and SIFT features set (the SIFT features were obtained as in [15]) to train the models.

Fig. 3 shows the simulation results on the CalTech5 dataset for different numbers of features. In general, it has been shown that PBIM outperforms BIM and SIFT in terms of accuracy for the most categories in the dataset. PBIM and BIM significantly outperform SIFT for the airplanes, faces, leaves, and cars; for the airplanes and leaves, PBIM is clearly superior to BIM, whereas for the faces and cars, PBIM can be competitive with BIM; PBIM did not achieve superior results in the motorcycles test.

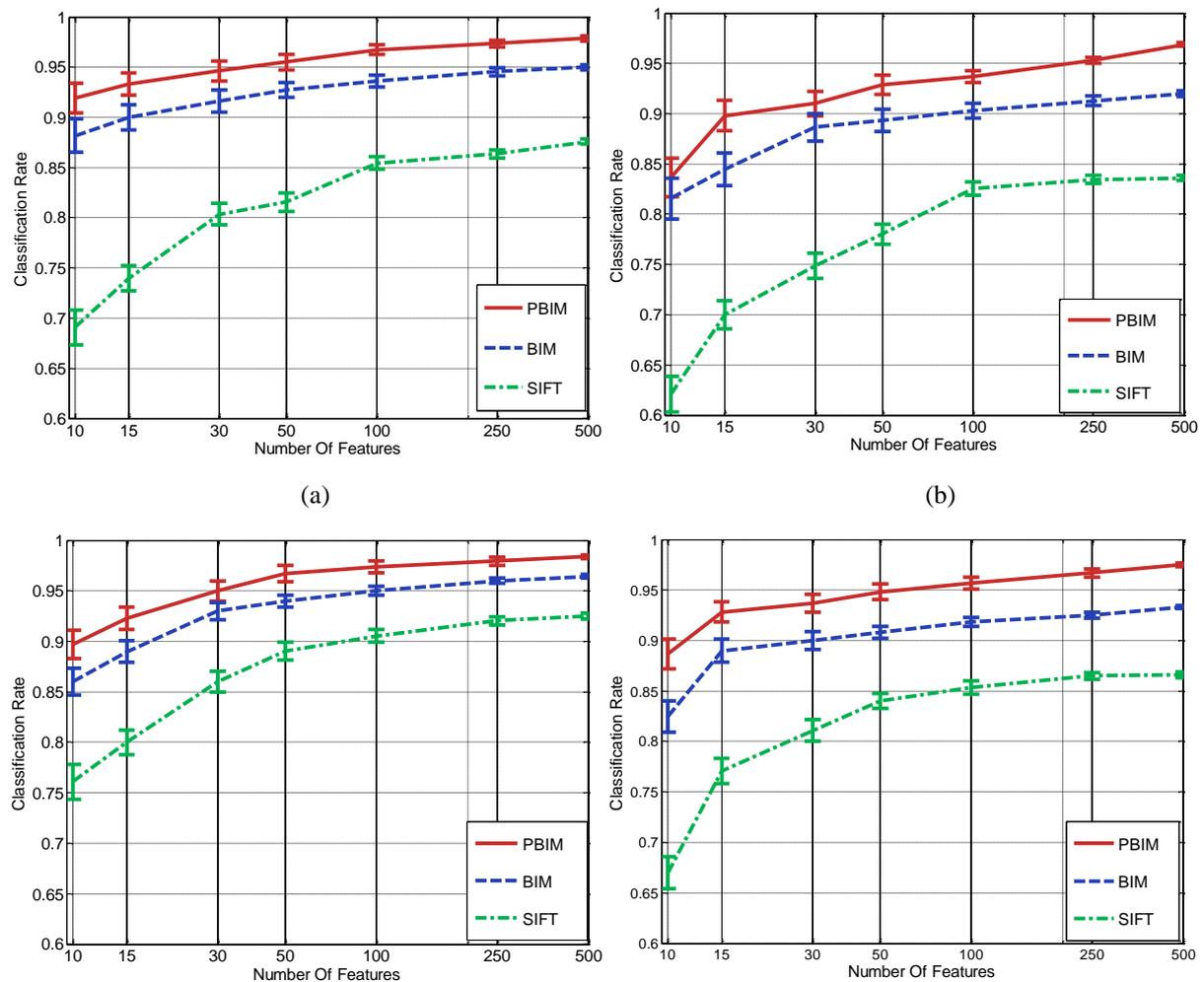

(c) (d)

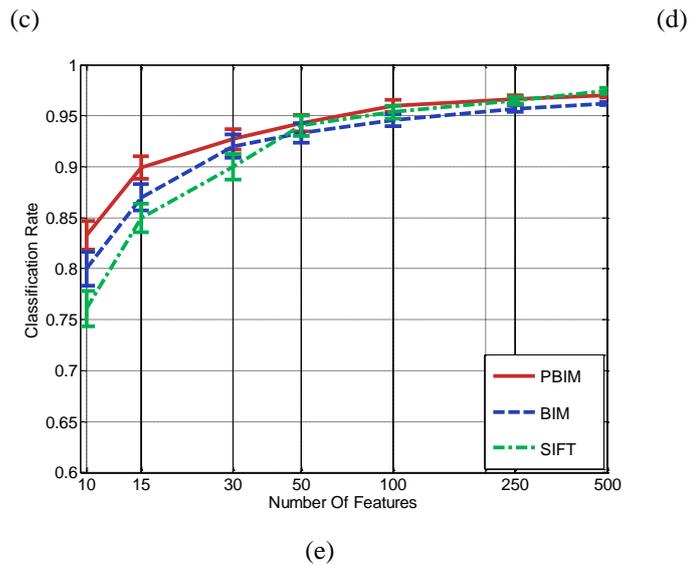

(e)

**Fig. 3.** Comparison of PBIM with SIFT and standard BIM on the CalTech5 database: (a)airplanes, (b) faces, (c) cars, (d) leaves, and (e) motorcycles.

### 4.2.2 TUD

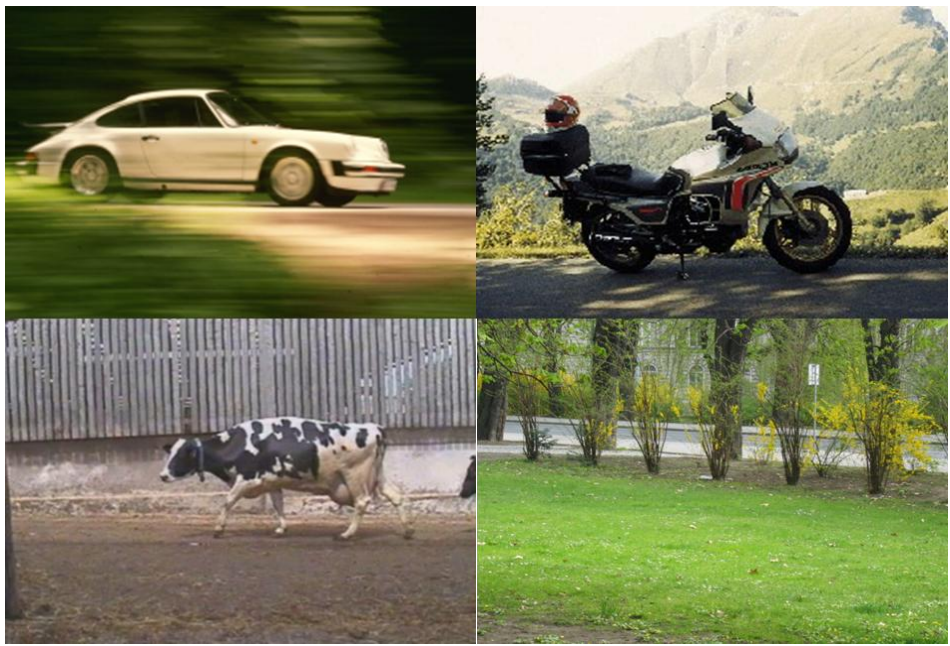

**Fig. 4.** Sampling images from the TUD dataset. The last image is a background image.

The TUD database (formerly the ETHZ database) contains side views of cars, motorcycles, and cows, as shown in Fig. 5. We evaluated the PBIM model, the conventional BIM that uses the random patch selection method, and a modified BIM model (MBIM) based on OGHM with random patches[16]. In addition, we also compared SIFT [5] and spatial pyramid matching using sparse coding (SPM) [30] in the experiment.

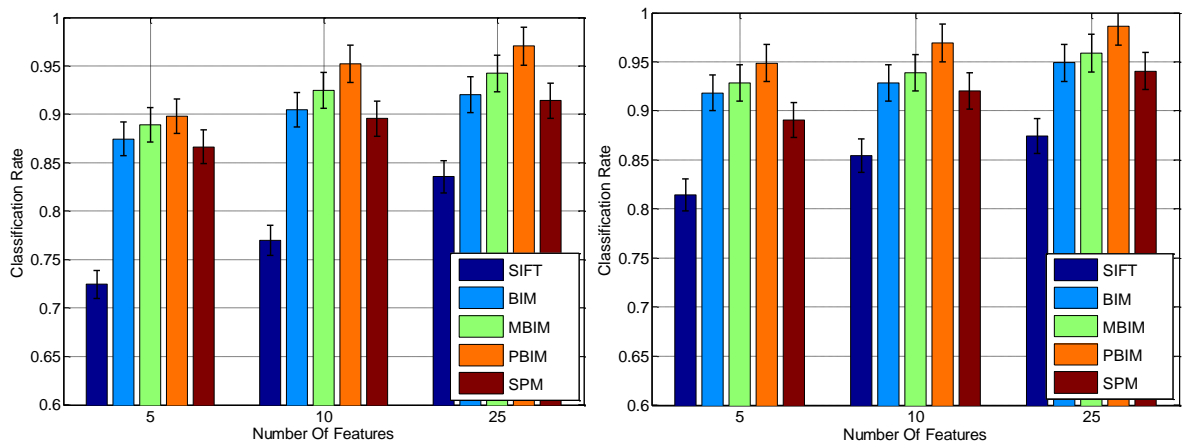

(a)　　　　　　　　　　　　　　　(b)

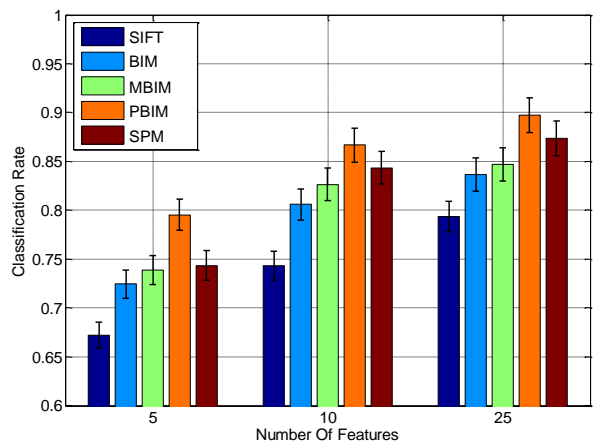

(c)

**Fig. 5.** Comparison of PBIM with standard BIM, MBIM, SIFT, and SPM on the TUD database: (a) cows, (b), cars and (c) motorcycles

To make the comparison at the feature level, we compared the scale and position-invariant C2 features of BIM models with the features produced by SIFT and SPM by passing them to a linear SVM that was trained to perform the object present/absent recognition task. We compared the classification rate for various numbers of features (5, 10, and 25). In the experiment, we randomly chose 15 images from each category of the TUD database as positive training images and 15 background images as the negative training set. For the tests, 50 images from each category of the TUD dataset and 50 images from backgrounds were randomly chosen as a test set. The results were generated from 10 independent trials. We report the mean and standard deviation of the classification across all classes.

Fig. 5 shows the simulation results on the TUD dataset for different numbers of features. In general, PBIM clearly outperforms SIFT, SPM, BIM, and MBIM in terms of accuracy for most of the categories in the dataset. In particular, PBIM significantly outperforms the other methods for the cars, and cows.

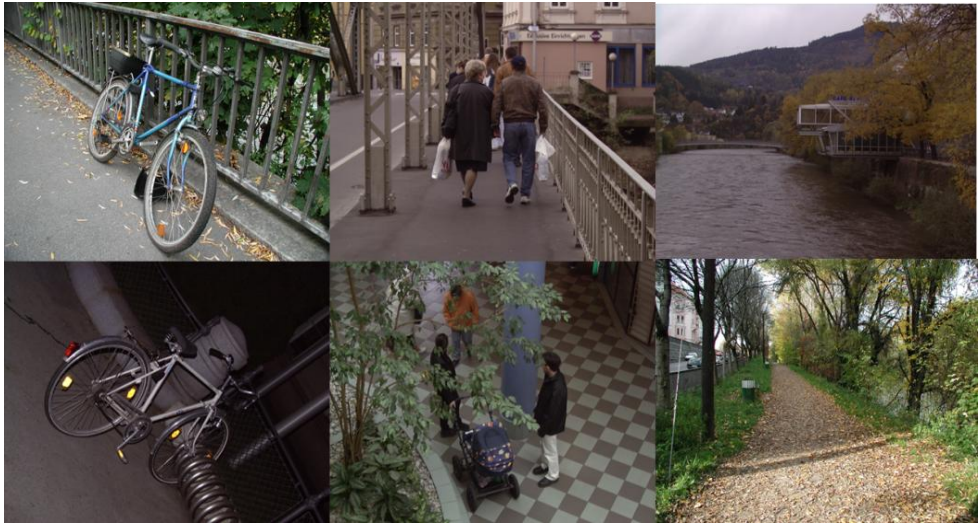

**Fig. 6.** Sampling images of GRAZ-01. From left to right, the categories are bikes, people and backgrounds.

### 4.2.3 GRAZ-01

GRAZ-01 [29] is a challenging dataset with high intra-class variability on highly cluttered backgrounds, containing persons, bikes, and backgrounds. The sampling images of the GRAZ-1 dataset are shown in Fig. 6. For the GRAZ-01 dataset, we followed the method presented in [29]: $100$ images (bike or person) and $100$ images (backgrounds) were randomly chosen as the training set; $50$ other images (bike or person) and other images (backgrounds) were chosen as the testing set. Fifteen hundred initial patches (features) were used for the experiment. We repeated the experiment 10 times and reported the averaged values of the test results. For effective evaluation of the PBIM model, we also tested the ROC and recall-precision (RP) curves and compared the performance of the proposed model with that of related approaches (i.e., Moment invariants, SIFT, Similarity-Measure-Segmentation (SM), modified Biologically Inspired Model (MBIM)) [29], [16]. The experimental GRAZ-01 dataset results are shown in Table 1.

Table 1 Performance Comparison of Several Approaches on GRAZ-01

| Method | Bikes | | Persons | |
| --- | --- | --- | --- | --- |
| | *EER | AUC | EER | AUC |
| Moment invariants | 73.5 | 76.5 | 63.0 | 68.7 |
| SIFT | 78.0 | 86.5 | 76.5 | 80.8 |
| SM | 83.5 | 89.6 | 56.5 | 59.1 |
| MBIM | 84.3 | 91.2 | 76.8 | 85.3 |
| PBIM | 85.5 | 94.3 | 86.9 | 92.7 |

* EER (detection rate at equal-error-rate of the ROC curve) and AUC (area under the ROC Curve)

Table 1 shows the ROC curves results: PBIM achieves the best performance in all cases. PBIM by far outperforms the moment invariants, SM, and SIFT approaches for both bikes and persons. The performance of PBIM are similar to that of the MBIM method at the bike cases ; however, PBIM significantly outperforms the MBIM method at the person recognition tasks. In general, our proposed model achieves competitive results.

## 5. Conclusion

In this article, we presented an OGHM based patch selection method (PSGHM), and extended the BIM model with the PSGHM method. The OGHM-based features have properties that are robust to in image distortions, including rotation. The proposed PBIM model increases the rotation invariance for local feature representation and refine the selected patches. PBIM provides a better balance between selective representation and invariance. Experiments on three different datasets demonstrated significant improvements as compared to the conventional BIM. Our work thus far has focused mainly on the low layers of BIM. Enhancing a deeper hierarchy of features will constitute our future work.

**Acknowledgment:** This work was supported by the National Science Foundation of China (Grant 61603389) and partially supported by National Natural Science Foundation of China (Grants 61502494, 61210009) and also by the Strategic Priority Research Program of the CAS (Grant XDB02080003).